\documentclass[10pt,twocolumn,letterpaper]{article}

\usepackage[pagenumbers]{cvpr} % To force page numbers, e.g. for an arXiv version
\usepackage{hyperref}
\usepackage{url}
\usepackage{graphicx}
\usepackage{wrapfig}
\usepackage{subcaption}
\usepackage{booktabs}       % professional-quality tables
\usepackage{multirow} 
\usepackage{amssymb}
\usepackage{pifont}
\usepackage{algorithm}
\usepackage{algorithmic}
\usepackage{pifont}

\usepackage{color} % (if used in text)

\usepackage{xcolor}
\usepackage{colortbl,booktabs}

\newcommand{\cmark}{\ding{51}} 
\newcommand{\xmark}{\ding{55}}

\hypersetup{
	colorlinks=true,
	citecolor=cyan ,
    linkcolor=red,
}
\def\paperID{13022} % *** Enter the Paper ID here
\def\confName{CVPR}
\def\confYear{2025}

\title{CacheQuant: Comprehensively Accelerated Diffusion Models}

\author{
Xuewen Liu$^{1,2}$, \; Zhikai Li$^{1,2\;*}$, \; Qingyi Gu$^{1}$ 
\thanks{Corresponding author: \{lizhikai2020, qingyi.gu\}@ia.ac.cn.} \\
$^1$Institute of Automation, Chinese Academy of Sciences\\
$^2$School of Artificial Intelligence, University of Chinese Academy of Sciences\\
}

\begin{document}
\maketitle
\begin{abstract}
Diffusion models have gradually gained prominence in the field of image synthesis, showcasing remarkable generative capabilities.
Nevertheless, the slow inference and complex networks, resulting from redundancy at both temporal and structural levels, hinder their low-latency applications in real-world scenarios.
Current acceleration methods for diffusion models focus separately on temporal and structural levels.
However, independent optimization at each level to further push the acceleration limits results in significant performance degradation.
On the other hand, integrating optimizations at both levels can compound the acceleration effects.
Unfortunately, we find that the optimizations at these two levels are not entirely orthogonal. Performing separate optimizations and then simply integrating them results in unsatisfactory performance.
To tackle this issue, we propose CacheQuant, a novel training-free paradigm that comprehensively accelerates diffusion models by jointly optimizing model caching and quantization techniques. 
Specifically, we employ a dynamic programming approach to determine the optimal cache schedule, in which the properties of caching and quantization are carefully considered to minimize errors.
Additionally, we propose decoupled error correction to further mitigate the coupled and accumulated errors step by step.
Experimental results show that CacheQuant achieves a 5.18$\times$ speedup and 4$\times$ compression for Stable Diffusion on MS-COCO, with only a 0.02 loss in CLIP score. Our \href{https://github.com/BienLuky/CacheQuant}{code} are open-sourced.

\end{abstract}    
\section{Introduction}
\label{sec:intro}
\begin{figure}[t]
    \centering
    \vspace{0.5cm}
    \includegraphics[width=1.0\linewidth]{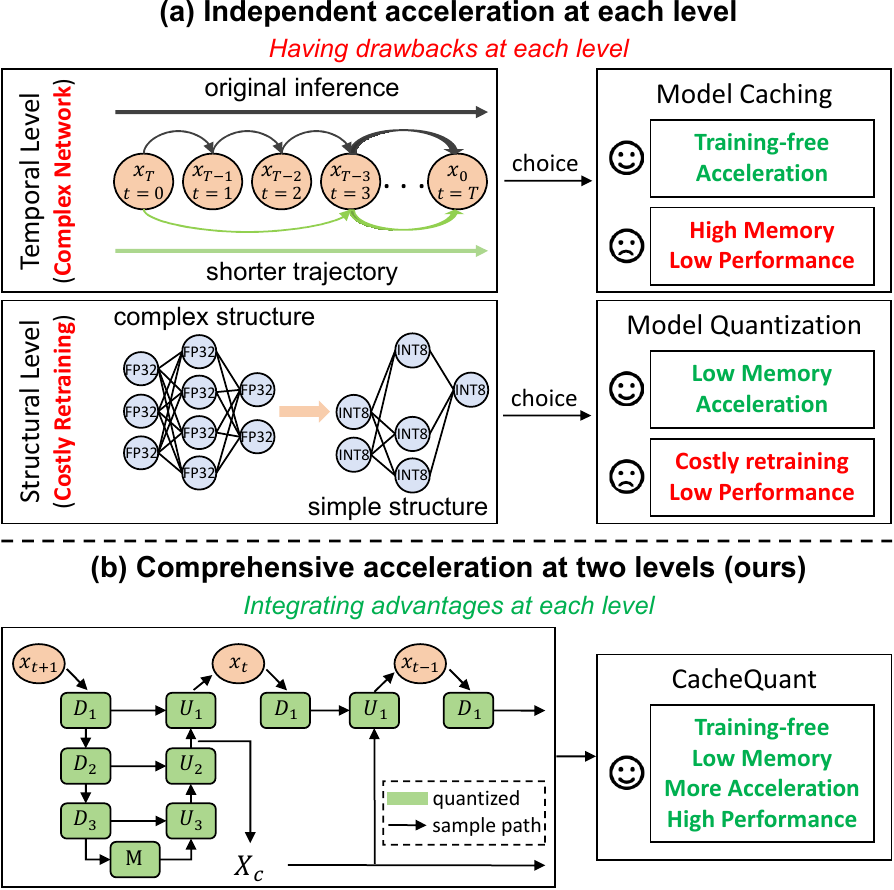}
    \caption{An overview of motivations. (a) The principles and properties of the traditional acceleration methods at each level. (b) Our approach integrates the advantages of model caching and quantization while eliminating their drawbacks, achieving comprehensive acceleration at two levels.}
    \label{fig:overview}
    \vspace{-0.3cm}
\end{figure}

Recently, diffusion models~\cite{dhariwal2021diffusion,ho2020denoising,song2019generative} with different frameworks, such as UNet~\cite{ronneberger2015u} and DiT~\cite{peebles2023scalable}, have come to dominate the field of image synthesis, exhibiting remarkable generative capabilities.
Numerous compelling applications have been implemented with diffusion models, including but not limited to image editing~\cite{avrahami2022blended,kawar2023imagic,meng2021sdedit}, image enhancing~\cite{li2022srdiff,saharia2022image,gao2023implicit}, image-to-image translation~\cite{choi2021ilvr,saharia2022palette,tumanyan2023plug}, text-to-image generation~\cite{nichol2021glide,ramesh2022hierarchical,saharia2022photorealistic,zhang2023adding} and text-to-3D generation~\cite{lin2023magic3d,luo2021diffusion,poole2022dreamfusion}.
Despite their appeal, the slow inference and complex networks, resulting from thousands of denoising iterations and billions of model parameters, pose significant challenges to deploy these models in real-world applications.
For instance, even on high-performance hardware A6000 GPU, a single inference of Stable Diffusion~\cite{rombach2022high} requires over a minute and consumes 16GB of memory.

To address the above challenges, the research community accelerates diffusion models primarily at two levels: the temporal level and the structural level.
For the former, existing methods~\cite{lu2022dpm,meng2023distillation,salimans2022progressive,song2020denoising,li2023faster} tackle the slow inference by shortening the denoising trajectory. In contrast, other methods~\cite{fang2023structural,castells2024ld,li2023q,liu2024enhanced,liu2024dilatequant} focus on simplifying the network structure to address the complex networks for the latter.
Although these methods have achieved significant results, each has its own drawbacks.
As shown in Figure~\ref{fig:overview}, temporal-level methods fail to reduce or even exacerbate the complexity of the networks, while structural-level methods require costly retraining processes.
Moreover, independent optimization at each level to push the acceleration limits, such as employing a shorter denoising path~\cite{ma2024deepcache} or further reducing model parameters~\cite{wang2024quest}, results in significant performance degradation.
Therefore, we seek to develop a comprehensive acceleration solution for diffusion models across both temporal and structural levels, aiming to integrating the advantages of each while eliminating their respective drawbacks. This allows us to push the acceleration boundaries further without compromising performance.

We start by analyzing the properties of methods at each level. 
At the temporal level, model caching~\cite{chen2024delta,wimbauer2024cache,selvaraju2024fora} utilize caching mechanisms to eliminate redundant computations at per step without any retraining, which preserve temporal continuity and maintain performance within equivalent computational budgets compared to other methods~\cite{nichol2021improved,song2020denoising,zhang2022gddim,liu2022pseudo,dockhorn2022genie}.
At the structural level, quantization-based methods~\cite{shang2023post,he2024ptqd} are more efficient in terms of training overhead and hardware friendly compared to other compression-based methods~\cite{li2024snapfusion,yang2023diffusion,liu2023oms,gu2023boot,sauer2023adversarial}.
Thus, we choice model caching and quantization to comprehensively accelerate diffusion models. Moreover, these two techniques exhibit a synergistic relationship: quantization reduces the memory usage increased by caching, while caching alleviates the quantization difficulties caused by temporal redundancy.

Based on the above analysis, theoretically, integrating optimized model caching with quantization methods can yield more substantial acceleration while maintaining controlled performance degradation.
However, in practice, we find that the optimizations at these two methods are not entirely orthogonal. Independently optimizing and then simply combining them results in unsatisfactory performance.
The underlying issue is that both caching and quantization introduce errors into the original models. These errors couple and accumulate iteratively, further exacerbating their impact on model performance and hindering the effective integration of optimization methods.
More specifically, if model quantization is applied directly to caching methods, the quantization error causes significant deviation in the denoising path of the cache. Conversely, if model caching is directly added to quantization methods, the caching error leads to substantial accumulation of quantization errors. In both cases, model performance degrades severely.

To this end, we introduce CacheQuant that solves the above issues by jointly optimizing model caching and quantization techniques.
Specifically, we propose Dynamic Programming Schedule (DPS) that models the design of the cache schedule as a dynamic programming problem, aiming to minimize the errors introduced by both caching and quantization.
Through optimization, the computational complexity of DPS is significantly reduced, requiring only 8 minutes for LDM on ImageNet. % with 250 time steps
To further mitigate the coupled and accumulated errors, we propose Decoupled Error Correction (DEC), which performs channel-wise correction separately for caching and quantization errors at each time step in a training-free manner.
Since the correction for quantization errors can be absorbed into weight quantization, EDC introduces only one additional matrix multiplication and addition during network inference.
To the best of our knowledge, this is the first work to investigate diffusion model acceleration at both the temporal and structural levels.
We also evaluate the acceleration capabilities of CacheQuant by deploying it on various hardware platforms (GPU, CPU, ARM).

In summary, we make the following contributions:
\begin{itemize}
    \item We introduce CacheQuant, a novel training-free paradigm that comprehensively accelerates diffusion models with different frameworks at both temporal and structural levels. Our method further pushes accelerated limits and maintains performance.
    \item CacheQuant minimizes the errors from caching and quantization through DPS and further mitigates these errors via DEC. It achieves a complementary advantage of model caching and quantization techniques by jointly optimizing them.
    \item We conduct experiments on diffusion models with UNet and DiT frameworks. Extensive experiments demonstrate that our approach outperforms traditional acceleration methods (solver, caching, distillation, pruning, quantization) in both speedup and performance. 
\end{itemize}

\begin{figure*}[t]
    \centering
    % \vspace{0.5cm}
    \includegraphics[width=1.0\linewidth]{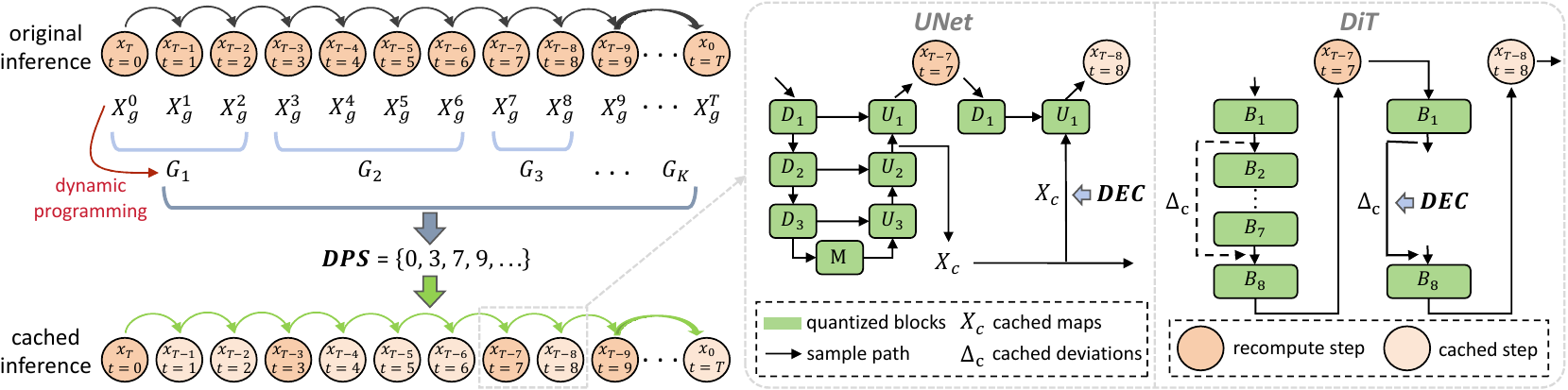}
    \caption{An overview of CacheQuant. \textbf{DPS} selects the optimal cache schedule and \textbf{DEC} mitigates the coupled and accumulated errors.}
    \label{fig:cachequant}
    \vspace{-0.2cm}
\end{figure*}
\section{Related Work}
\label{sec:related}
Diffusion models have gradually surpassed GANs~\cite{arjovsky2017wasserstein,goodfellow2014generative} and VAEs~\cite{higgins2017beta,kingma2013auto}, emerging as the dominant approach in image generation. 
However, slow inference and complex networks hinder their low-latency applications in real-world scenarios. 
Current research focuses on two main levels to accelerate diffusion models.

%-------------------------------------------------------------------------
\vspace{-0.2cm}
\paragraph{Temporal-level Acceleration}
methods focus on shortening the sampling trajectory.
Some approaches adjust variance schedule~\cite{nichol2021improved} or modify denoising equations~\cite{song2020denoising,zhang2022gddim} to remove certain steps entirely.
% DDIM~\cite{song2020denoising} explore a non-Markovian process to reduce the entire steps. 
Studies further dive into the fast solver of SDE~\cite{liu2022pseudo,dockhorn2022genie} or ODE~\cite{lu2022dpm, lu2022dpm1} to create efficient sampling steps.
Others~\cite{zheng2023fast,shih2024parallel,li2024distrifusion} conduct parallel sampling to speed up inference.
In contrast, cache-based methods~\cite{ma2024deepcache,chen2024delta,wimbauer2024cache} reduce the inference path at each step by caching the output of block.

%-------------------------------------------------------------------------
\vspace{-0.2cm}
\paragraph{Structural-level Acceleration}
methods concentrate on simplifying the network architecture.
Previous studies redesign lightweight network~\cite{li2024snapfusion} or incorporate frequency priors into model design~\cite{yang2023diffusion}.
OMS-DPM~\cite{liu2023oms} creates a diffusion model zoo to select different models at various steps.
Some methods~\cite{salimans2022progressive,meng2023distillation,luo2023latent} simplify model architecture with distillation technology.
On the other hand, pruning-based methods~\cite{zhang2024laptop,fang2023structural,castells2024ld} reduce the number of model parameters, while quantization-based approaches~\cite{liu2024enhanced,liu2024dilatequant,li2023q,li2023vit,li2022psaqvit} achieve model compression by utilizing lower bit-width representations.

\section{Preliminary}
In the following, we present the two key techniques employed in our work.
\vspace{-0.2cm}
\paragraph{Model Caching}
accelerates inference by storing intermediate network outputs. 
For diffusion models with temporal networks, this technique leverages the inherent similarity of feature maps between adjacent denoising steps to eliminate temporal computational redundancies.
For example, we cache the output activation $X_g^t$ of block at step $t$ as $X^t_{c}$. When inferring at step $t+1$, $X^t_{c}$ is reused in place of the ground truth $X^{t+1}_{g}$, thereby eliminating the computations of $X^{t+1}_{g}$.
Existing methods implement caching at various network layers.
Deepcache~\cite{ma2024deepcache} and Faster Diffusion~\cite{li2023faster} cache the output feature maps of upsampling blocks and UNet encoder, respectively.
Block Caching~\cite{wimbauer2024cache} further adaptively caches all blocks.
$\Delta$-DiT~\cite{chen2024delta} selectively caches blocks based on their impact at different denoising stages.
Besides, this mechanism can be extended to cover more steps, with the cached features $X^t_{c}$ calculated once and reused in the consecutive $N-1$ steps:
\begin{equation}
  X^t_{c} \to X^{t+1}_{g} \Rightarrow X^t_{c} \to \{X^{t+1}_{g}, X^{t+2}_{g}, ... , X^{t+N}_{g}\} 
\end{equation}
Determining the cache schedule, i.e., where to recompute cached features, directly impacts model performance. For instance, in a diffusion model with $T$ steps, when the cache frequency $N$ is fixed, a uniform cache schedule is represented by $\{0, N, 2N, ... , (T/N-1)N\}$, and the corresponding cached features are $\{X^{0}_{c}, X^{N}_{c}, X^{2N}_{c}, ... , X^{(T/N-1)N}_{c}\}$.
To reduce the errors introduced by caching, previous methods have developed various cache schedules.
\cite{ma2024deepcache,wimbauer2024cache,chen2024delta} determine the schedule by conducting experiments and tuning hyperparameters, while~\cite{li2023faster} directly specifies the schedule manually.
In this work, as shown in Figure~\ref{fig:cachequant}, for the UNet framework, we cache the outputs of a single upsampling block as the cached features $X_c$, similar to the approach used in DeepCache~\cite{ma2024deepcache}.
For the DiT framework, we cache the deviations $\Delta_c$ between two blocks as the cached features, similar to $\Delta$-DiT~\cite{chen2024delta}.
We model the selection of schedule as a dynamic programming problem and our goal is to minimize the errors introduced by caching and quantization, thereby achieving the optimal schedule.

\vspace{-0.2cm}
\paragraph{Model Quantization}
represents model parameters and activations with low-precision integer values, compressing model size and accelerating inference. Given a floating-point vector $\mathbf{x}$, it can be uniformly quantized as follows:
\begin{equation}
    \hat{\mathbf{x}}=clip \left( \left\lfloor {\mathbf{x}} / {s}\right\rceil+z, 0, 2^b-1 \right)
\end{equation}
where $\hat{\mathbf{x}}$ is the quantized value, scale factor $s$ and zero point $z$ are quantization parameters, $\left\lfloor \cdot \right\rceil$ denotes rounding function, and the bit-width $b$ determines the range of clipping function $clip(\cdot )$.
Depending on whether fine-tuning of the model is necessitated, this technique
can be categorized into two approaches: post-training quantization (PTQ) and quantization-aware training (QAT).
Initial PTQ methods~\cite{nagel2019data,li2023repq} calibrate the quantization parameters in a training-free manner using a small calibration set. Subsequently, reconstruction-based methods~\cite{wei2022qdrop,nagel2020up,li2024repquant} employ backpropagation to optimize quantization parameters.
On the other hand, QAT methods~\cite{gong2019differentiable,nagel2022overcoming} entail fine-tuning the model weights on the original dataset.
While this approach preserves performance, it requires significant time cost and computational resources.
Notably, all existing quantization methods for diffusion models are either reconstruction-based~\cite{liu2024enhanced,li2023q} or fine-tuning-based~\cite{liu2024dilatequant,he2023efficientdm}. In stark contrast, we propose a PTQ correction strategy to mitigate errors, preserving the advantages of being training-free.

\section{CacheQuant}
In this section, we introduce \textbf{CacheQuant}, a novel training-free paradigm that jointly optimizes caching and quantization techniques to comprehensively accelerate diffusion models.
We start by analyzing the challenges of comprehensive acceleration in Sec~\ref{sec:4.1}, followed by our proposed methods to address these challenges in Sec~\ref{sec:4.2} and Sec~\ref{sec:4.3}.
The overview of CacheQuant is shown in Figure~\ref{fig:cachequant}.

\subsection{Challenges of Comprehensive Acceleration}\label{sec:4.1}
\begin{figure}[t]
    \centering
    % \vspace{-0.2cm}
    \includegraphics[width=1.0\linewidth]{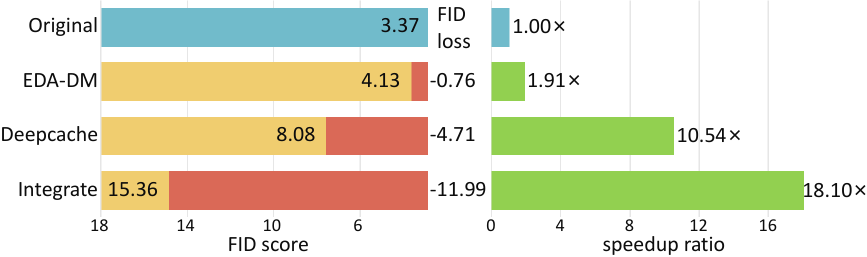}
    \caption{Performance and acceleration of different optimization strategies. EDA-DM and Deepcache are optimization methods for model quantization and caching, respectively. }
    \label{fig:issue}
    \vspace{-0.3cm}
\end{figure}

\begin{figure}[t]
    \centering
    % \vspace{-0.2cm}
    \includegraphics[width=1.0\linewidth]{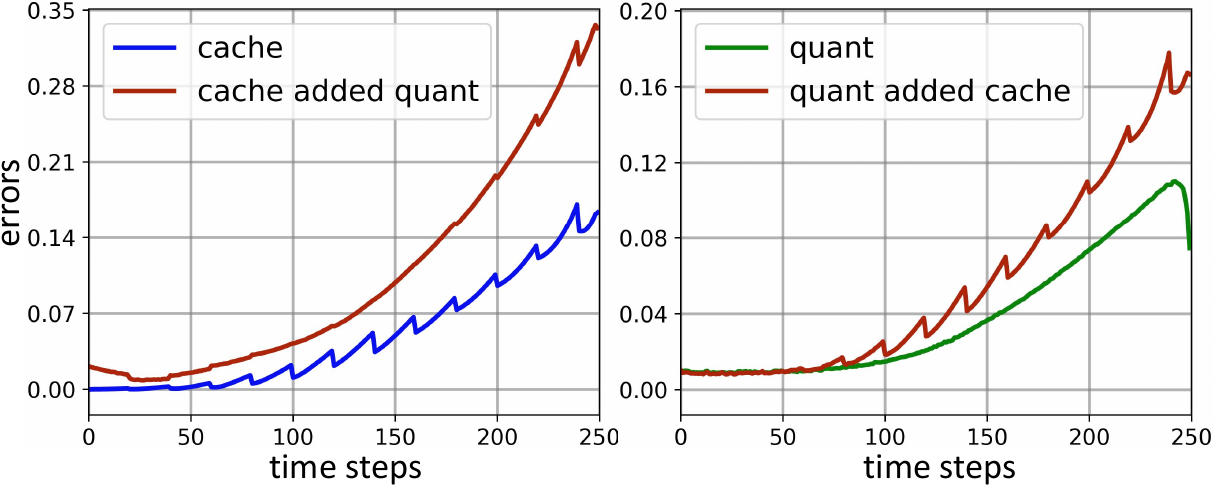}
    \caption{Output errors of network at each time step.}
    \label{fig:error}
    \vspace{-0.3cm}
\end{figure}

Leveraging model caching and quantization enables comprehensive acceleration of diffusion models.
Unfortunately, we find that, although independently optimizing and then simply integrating these two methods yields more noticeable acceleration, the model performance remains far from satisfactory.
To analyze the above issues, we conduct experiments for LDM on ImageNet.
As shown in Figure~\ref{fig:issue}, when the original model is independently optimized through model quantization and caching, the FID score drops 0.76 and 4.71, respectively.
However, simply integrating the two optimizations results in an FID loss of 11.99.
The underlying issue is that both caching and quantization inherently introduce errors into the original models. These errors couple and accumulate iteratively, further exacerbating their impact on model performance and hindering the effective combination of optimization methods.
As shown in Figure~\ref{fig:error}, if model quantization is directly added to caching methods, the quantization error causes significant deviation in the denoising path of the cache. Conversely, if model caching is applied directly to quantization methods, the caching error leads to substantial accumulation of quantization errors. 
This suggests that the optimizations of these two methods are not entirely orthogonal, highlighting the need for joint optimization.

\subsection{Dynamic Programming Schedule}\label{sec:4.2}
We illustrate our method with the UNet framework as an example.
To minimize errors, we analyze the feature maps $X = \{X^0_g, X^1_g, ..., X^{T-1}_g\}$ at all steps to guide the selection of the cache schedule, reframing the problem as one of grouping ordered samples.

For a diffusion model with $T$ steps and a cache frequency of $N$, all feature maps are divided into $K = T/N$ groups, forming a grouping set $G = \{G_1, G_2, ..., G_K\}$.
Time steps within the same group share the same cached features.
To achieve optimal grouping, we propose \textbf{D}ynamic \textbf{P}rogramming \textbf{S}chedule (\textbf{DPS}):

First, we consider two constraints: 1) Each feature map belongs to exactly one group, ensuring that no step is duplicated or omitted; 2) The order of the feature maps within each group must remain unchanged to preserve the temporal consistency of the denoising process.
Specified as:
\begin{equation}
    \begin{aligned}
        G_1 &= \{X^0_g, X^1_g, ..., X^{s_1-1}_g\}, \\
        G_2 &= \{X^{s_1}_g, X^{s_1+1}_g, ..., X^{s_2-1}_g\}, \\
        & \; ... \\
        G_K &= \{X^{s_{K-1}}_g, X^{s_{K-1}+1}_g, ..., X^{T-1}_g\}
    \end{aligned}
\end{equation}
where the time step of the first element $X^{s_{*}}_g$ of each group denotes the dividing point, which forms the cache schedule.

\begin{algorithm}[t]
    \caption{: Dynamic Programming Schedule}
    \label{alg:DPS}
    \textbf{Input}: all steps $T$, cache frequency $N$ \\
    \textbf{Output}: optimal schedule $DPS$ 
    \begin{algorithmic}
    \STATE $K = T / N$ \hspace*{\fill}{{\color{gray}$\triangleright$ init number of groups}}
    \STATE $M = S = 0_{T \times K}$ \hspace*{\fill}{{\color{gray}$\triangleright$ init loss $M$ and dividing point $S$}}
    \FOR{$t=1$ to $T$} 
        \STATE $M(t, 1) = D(1, t)$ \hspace*{\fill}{{\color{gray}$\triangleright$ calculate boundary conditions}}
        \STATE $S(t, 1) = t $
    \ENDFOR
    \FOR{$k=1$ to $K$}
        \FOR{$t=k$ to $T$}
            \STATE clear $L$ = [ ]
            \FOR{$s=k$ to $t$}
                \STATE limit {\color[HTML]{1E90FF}\text{$\frac{1}{2}N \leq t-s \leq 2N$}} \hspace*{\fill}{{\color{gray}$\triangleright$ optimization limits}}
                \STATE $L[b(t, k)] = M(s-1, k-1) + D(s, t)$
                \STATE append $L[b(t, k)]$ to $L$
            \ENDFOR
            \STATE $M(t, k) = \min (L)$ \hspace*{\fill}{{\color{gray}$\triangleright$ store mininum loss}}
            \STATE $S(t, k) = argmin _{s} (L) $ \hspace*{\fill}{{\color{gray}$\triangleright$ store dividing point}}
        \ENDFOR
    \ENDFOR
    \STATE t = T
    \FOR{$k=K$ to $1$}
        \STATE $s_k = S(t, k)$  \hspace*{\fill}{{\color{gray}$\triangleright$ dividing point for $k$-th group}}
        \STATE append $s_k$ to $DPS$ \hspace*{\fill}{{\color{gray}$\triangleright$ store for optimal schedule}}
        \STATE $t = s_k - 1$ \hspace*{\fill}{{\color{gray}$\triangleright$ number of remaining steps}}
    \ENDFOR
    \end{algorithmic}
\end{algorithm}

Second, we define the intra-group error as $D_k(i,j)$, which represents the error introduced by caching and quantization when steps $i$ to $j$ are assigned to the $k$-{th} group.
Notably, since $X^i_g$ is cached and replaces $\{ X^{i+1}_g, ..., X^j_g \}$, the error is calculated by sequentially comparing $X^i_g$ with $\{ X^{i+1}_g, ..., X^j_g \}$ and summing the resulting differences.
Additionally, quantization error arises from the absolute numerical differences between feature maps, and is therefore measured using the $L1$ norm.
% We demonstrate the validity of two errors calculations in Appendix. 
Consequently, the mathematical formulation of $D_k(i,j)$ is as follows:
\begin{equation}
    \begin{aligned}
        D_k(i,j) = \sum_{t=i+1}^{j} \| X^{i}_g - X^{t}_g \|_1
    \end{aligned}
\end{equation}

Third, we denote the partitioning of $T$ steps into $K$ groups as $b(T, K)$. The grouping loss function is defined as $L[b(T, K)] = \sum_{k=1}^{K} D_k(i,j)$.
The solution for $K$-th optimal group can be expressed as:
\begin{align}
    L[b(T, K)] &=L[b(s-1, K-1)]+D(s, T) \\
    M(T, K) &=\min _{K \leq s \leq T} L[b(T, K)]
\end{align}
where $s$ denotes the dividing point, ${K \leq s \leq T}$ ensures that each feature map belongs to exactly one group, $M(T, K)$ minimizes the grouping loss to obtain the $K$-th optimal group $G_K=\{X^{s}_g, X^{s+1}_g, ..., X^{T-1}_g\}$.
Briefly, the above formula can be reformulated as:
\begin{align}
    M(T, K) &=\min _{K \leq s \leq T} \{ M(s-1, K-1) + D(s, T) \}
\end{align}
As can be seen, the $K$-th optimal group is based on the assignment of the $s-1$ feature maps to $K-1$ optimal groups.
Thus, all optimal groups can be iteratively solved based on the boundary conditions $M(t, 1)$. 
The workflow of DPS is shown in Algorithm~\ref{alg:DPS}.

However, due to the nested loops, the computation of DPS is complex, resulting in slow convergence. 
We consider practical grouping factors, optimizing the group length to no more than $2N$ and no less than $\frac{1}{2}N$. This significantly reduces the computational complexity of DPS.
For instance, the solution time for LDM with 250 steps on ImageNet is reduced from 4 hours to 8 minutes.
Finally, DPS efficiently obtains the optimal schedule that minimizes both caching and quantization errors.

\begin{figure*}[t]
    \centering
    % \vspace{0.5cm}
    \includegraphics[width=1.0\linewidth]{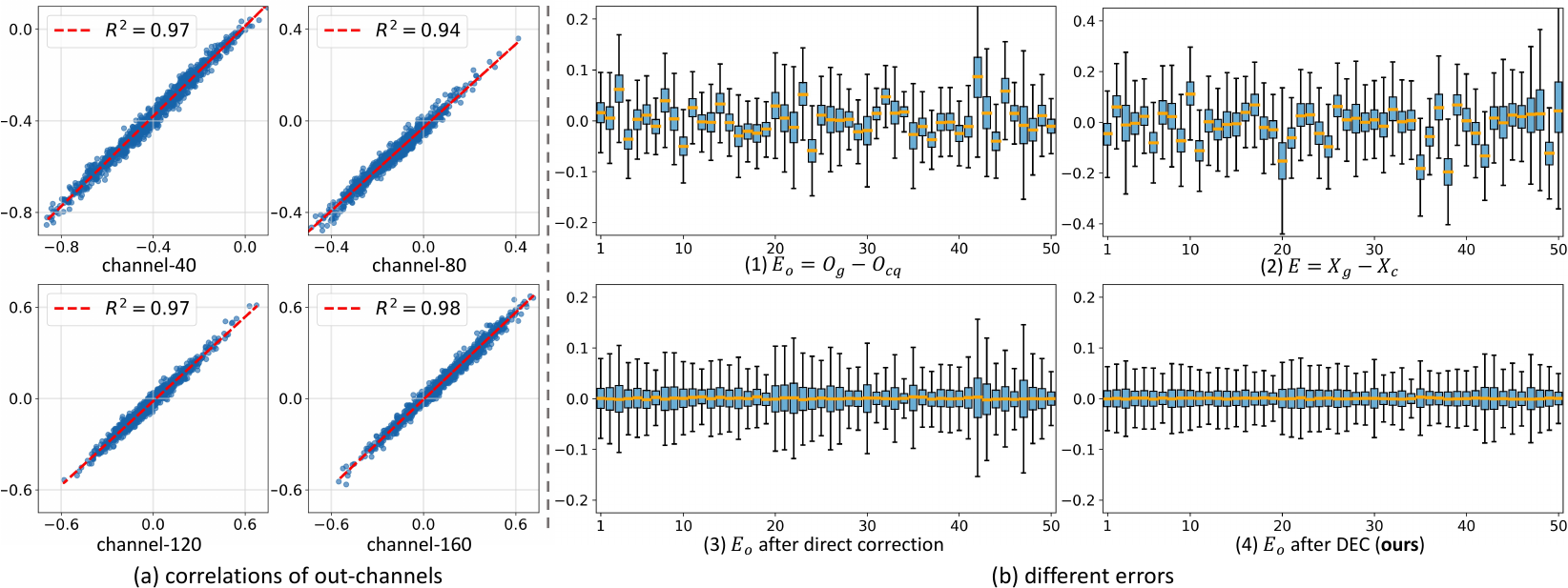}
    \caption{(a) Correlations between the different out-channels of $O_g$ and $O_{cq}$. (b) Box plots visualize the mean and variance of different errors. Data comes from steps $t=192$ and $t=210$ for LDM on ImageNet, which are assigned to the same group by the DPS.}
    \label{fig:dec}
    \vspace{-0.2cm}
\end{figure*}
\subsection{Decoupled Error Correction}\label{sec:4.3}
To further mitigate the coupled and accumulated errors while maintaining acceleration efficiency, we explore a training-free solution.
We begin by analyzing the outputs of block receiving cached features under different conditions:
\begin{align}\label{eq:O}
    O_g = X_g W_g \;,\; O_c = X_c W_g \;,\; O_{cq} = X_{cq} W_q 
\end{align}
where $O \in \mathbb{R}^{B\times C^o}$, $X \in \mathbb{R}^{B\times C^i}$, and $W \in \mathbb{R}^{C^i\times C^o}$ denote the output, activation, and weight, respectively. The $B$, $C^i$ and $C^o$ denote the batch size, in-channel dimension, and out-channel dimension, respectively.
The subscripts $g$, $c$, and $q$ represent the different conditions: ground truth, cached, and quantized, respectively.
We observe a strong correlation in channel-wise granularity between $O_g$ and $O_{cq}$, as shown in Figure~\ref{fig:dec}(a).
Therefore, we can calculate correction parameters along the out-channel dimension for $O_{cq}$, aiming to reduce their error relative to $O_g$. 
And we correct at each step to alleviate accumulated errors.
The corrected formula is as follows:
\begin{align}\label{eq:cor}
    O_g = a \cdot O_{cq} + b
\end{align}
where $a \in \mathbb{R}^{C^o}$ and $b \in \mathbb{R}^{C^o}$ are correction parameters. 
We solve them using the least squares method. For instance, the correction parameters for $k$-th channel are as follow:
\begin{equation}\label{eq:ab}
    \begin{aligned}
        a_k &= \frac{\operatorname{Cov}(O_{cq(:, k)}, O_{g(:, k)})}{\operatorname{Var}(O_{cq(:, k)})} \\
        b_k &= \bar{O}_{g(:, k)} - a_k * \bar{O}_{cq(:, k)}
    \end{aligned}
\end{equation}
Here, $\bar{O}_{g(:, k)}$ and $\bar{O}_{cq(:, k)}$ represent the mean of the 
$k$-th out-channel.
When adjusting $O_{cq}$ using the correction parameters, although the mean error is eliminated, the variance of the error remains large, resulting in ineffective correction, as shown in Figure~\ref{fig:dec}(b)(1) and (3).
The underlying issue is that directly correcting $O_{cq}$ cannot efficiently eliminate caching errors, as these errors fundamentally arise from the difference between $X_g$ and $X_c$.

To address this, we propose \textbf{D}ecoupled \textbf{E}rror \textbf{C}orrection (\textbf{DEC}) that decouples error $E_o$ introduced by caching and quantization into cache error $E_c$ and quantization error $E_q$:
% \begin{align}
%     E_o &= X_g W_g - X_{cq} W_q = O_g - O_{cq} \\
%     E_c &= X_g W_g - X_c W_g = O_g - O_c \\
%     E_q &= X_c W_g - X_{cq} W_q = O_c - O_{cq}
% \end{align}
\begin{equation}
    \begin{aligned}
        E_o &= X_g W_g - X_{cq} W_q = O_g - O_{cq} \\
        E_c &= X_g W_g - X_c W_g = O_g - O_c \\
        E_q &= X_c W_g - X_{cq} W_q = O_c - O_{cq}
    \end{aligned}
\end{equation}
Similar to Eq.~\ref{eq:cor}, we correct $X_c$ to reduce $E_c$ and correct $O_{cq}$ to reduce $E_q$:
\begin{equation}
    \begin{aligned}
        X_g &= a_1 \cdot X_c + b_1 \\
        O_c &= a_2 \cdot O_{cq} + b_2
    \end{aligned}
\end{equation}
where the correction parameters $(a_1, b_1) \in \mathbb{R}^{C^i}$, $(a_2, b_2) \in \mathbb{R}^{C^o}$ are solved like Eq.~\ref{eq:ab}.
Experimental results demonstrate that DEC not only eliminates the mean error but also efficiently reduces error variance (shown in Figure~\ref{fig:dec}(b)(4)), significantly improving performance. 
For instance, compared to direct correction, DEC enhances the FID score for LDM on ImageNet by 0.91.

We also provide a theoretical proof that DEC outperforms direct correction. 
Through equivalent transformations (please see Appendix~\ref{app:Xcq} for details), the two correction methods express $O_{cq}$ as:
\begin{align}
    O_{cq} = X_{cq} W_q &= \frac{X_g W_g}{a} - \frac{b}{a} \\
    O_{cq} = X_{cq} W_q &= \frac{X_g}{a_1} \cdot \frac{W_g}{a_2} - \frac{b_1}{a_1} \cdot \frac{W_g}{a_2} - \frac{b_2}{a_2}
\end{align}
As can be seen, compared to direct correction on the out-channels, DEC adjusts the mean and variance across both in-channels and out-channels. The two expressions are equivalent when assuming $a_1 = \mathbf{1}$ and $b_1 = \mathbf{0}$, which implies that the mean error between $X_g$ and $X_c$ is zero and the variance is negligible. However, as shown in Figure~\ref{fig:dec}(b)(2), this assumption clearly does not hold, making DEC the more reasonable approach.
Additionally, by incorporating $(a_2, b_2)$ into weight quantization, DEC introduces only one additional matrix multiplication and addition during network inference.

\section{Experiment}\label{sec:exp}
\subsection{Experimental Setup}
\paragraph{Models, Datasets, and Metrics.}
To demonstrate the effectiveness of our method, we conduct evaluations on DDPM, LDM, and Stable Diffusion~\cite{song2020denoising,rombach2022high} with UNet framework and DiT-XL/2~\cite{Peebles2022DiT} with DiT framework. 
We present experimental results on six commonly used datasets: CIFAR-10, LSUN-Bedroom, LSUN-Church, ImageNet, MS-COCO, and PartiPrompt~\cite{krizhevsky2009learning,yu2015lsun,deng2009imagenet,lin2014microsoft,yu2022partiprompt}. 
Following previous works~\cite{liu2024dilatequant,ma2024deepcache,chen2024delta,wu2024ptq4dit}, we utilize 5k validation set from MS-COCO and 1.63k captions from PartiPrompt as prompts for Stable Diffusion, and generate 10k images for DiT-XL/2. 
For other tasks, we generate 50k images to assess the generation quality.
The evaluation metrics include FID, IS, and CLIP Score (on ViT-g/14)~\cite{heusel2017fid,hessel2021clipscore,salimans2016IS}. 
Besides, we employ Bops ($Bops=MACs \times b_w \times b_x$), Speed Up (on GPU), and Model Size (MB) to visualize acceleration and compression performance.

\begin{table}[t]\scriptsize %\footnotesize %\tiny %\small
    \centering
    \caption{Unconditional generation quality on CIFAR-10, LSUN-Church, and LSUN-Bedroom using DDPM, LDM-8, and LDM-4, respectively. The notion `WxAy' is employed to represent the bit-widths of weights `W' and activations `A'. }
    \label{tab:cifar}
    \setlength{\tabcolsep}{1.10mm}
    
\begin{tabular}{c l | c c c c | c }
  \toprule
  \bf Dataset & \bf Method  & \bf Bops $\downarrow$ & \bf Speed $\uparrow$ & \bf Size $\downarrow$ & \bf Retrain & \bf FID $\downarrow$ \\
  \midrule
  \multirow{7}{1.10cm}{\centering \\CIFAR \\ 32 $\times$ 32 \\T = 100} & DDPM & 6.21T & 1.00$\times$ & 143.0 & \xmark & 4.19 \\
  \cmidrule(r){2-7}
  \multirow{7}{*}{} & Deepcache-N=3 & 3.62T & 1.61$\times$ & 1.00$\times$ & \xmark & 4.70 \\
  \multirow{7}{*}{} & Ours-N=3 (W8A8) & 0.23T & 3.57$\times$ & 3.98$\times$ & \xmark & 4.61 \\
  \cmidrule(r){2-7}
  \multirow{7}{*}{} & Deepcache-N=5 & 3.08T & 1.85$\times$ & 1.00$\times$ & \xmark & 5.73 \\
  \multirow{7}{*}{} & Ours-N=5 (W8A8) & 0.19T & 4.11$\times$ & 3.98$\times$ & \xmark & 5.28 \\
  \cmidrule(r){2-7}
  \multirow{7}{*}{} & Deepcache-N=10 & 2.69T & 2.07$\times$ & 1.00$\times$ & \xmark & 9.74 \\
  \multirow{7}{*}{} & Ours-N=10 (W8A8) & 0.17T & 4.62$\times$ & 3.98$\times$ & \xmark & 8.19 \\
  \midrule
  \multirow{7}{1.10cm}{\centering \\LSUN-Church \\ 256 $\times$ 256\\T = 100\\eta=0.0} & LDM-8 & 19.10T & 1.00$\times$ & 1514.5 & \xmark & 3.99 \\
  \cmidrule(r){2-7}
  \multirow{7}{*}{} & Deepcache-N=2 & 10.07T & 1.86$\times$ & 1.00$\times$ & \xmark & 4.43 \\
  \multirow{7}{*}{} & Ours-N=2 (W8A8) & 0.63T & 3.10$\times$ & 3.99$\times$ & \xmark & 3.52 \\
  \cmidrule(r){2-7}
  \multirow{7}{*}{} & Deepcache-N=3 & 7.18T & 2.54$\times$ & 1.00$\times$ & \xmark & 5.10 \\
  \multirow{7}{*}{} & Ours-N=3 (W8A8) & 0.45T & 4.14$\times$ & 3.99$\times$ & \xmark & 3.66 \\
  \cmidrule(r){2-7}
  \multirow{7}{*}{} & Deepcache-N=5 & 4.65T & 3.67$\times$ & 1.00$\times$ & \xmark & 6.74 \\
  \multirow{7}{*}{} & Ours-N=5 (W8A8) & 0.29T & 5.98$\times$ & 3.99$\times$ & \xmark & 3.71 \\
  \midrule
  \multirow{7}{1.10cm}{\centering \\LSUN-Bedroom \\ 256 $\times$ 256\\T = 100\\eta=0.0} & LDM-4 & 98.36T & 1.00$\times$ & 1317.4 & \xmark & 10.49 \\
  \cmidrule(r){2-7}
  \multirow{7}{*}{} & Deepcache-N=2 & 52.23T & 1.79$\times$ & 1.00$\times$ & \xmark & 11.21 \\
  \multirow{7}{*}{} & Ours-N=2 (W8A8) & 3.26T & 3.05$\times$ & 3.99$\times$ & \xmark & 8.85 \\
  \cmidrule(r){2-7}
  \multirow{7}{*}{} & Deepcache-N=3 & 37.49T & 2.68$\times$ & 1.00$\times$ & \xmark & 11.86 \\
  \multirow{7}{*}{} & Ours-N=3 (W8A8) & 2.34T & 4.72$\times$ & 3.99$\times$ & \xmark & 9.27 \\
  \cmidrule(r){2-7}
  \multirow{7}{*}{} & Deepcache-N=5 & 24.59T & 4.08$\times$ & 1.00$\times$ & \xmark & 14.28 \\
  \multirow{7}{*}{} & Ours-N=5 (W8A8) & 1.54T & 7.06$\times$ & 3.99$\times$ & \xmark & 10.29 \\
  \bottomrule
\end{tabular}
    \vspace{-0.3cm}
\end{table}

\vspace{-0.2cm}
\paragraph{Caching and Quantization Settings.}
Our method uses Deepcache~\cite{ma2024deepcache}, $\Delta$-DiT~\cite{chen2024delta}, and EDA-DM~\cite{liu2024enhanced} as the baseline. 
We select the last 3/1/1-th blocks as cached blocks for DDPM, LDM, and Stable Diffusion models, respectively, and maintain Middle Blocks ($I=7$ and $N_c=14$ in~\cite{chen2024delta}) as cached object for DiT-XL/2. 
For model quantization, we utilize the temporal quantizer from~\cite{liu2024dilatequant} to quantize all layers, with channel-wise quantization for weights and layer-wise quantization for activations, as this is the common practice. 
Additionally, CacheQuant seamlessly integrates with quantization reconstruction to enhance performance. 

\subsection{Comparison with Temporal-level Methods}
The mainstream temporal-level acceleration methods for diffusion models include model caching and fast solvers.
We first compare CacheQuant with cache-based methods (Deepcache~\cite{ma2024deepcache}, $\Delta$-DiT~\cite{chen2024delta}), as reported in Table~\ref{tab:imagenet} and~\ref{tab:imagenet1}. 
Our method achieves comparable or even superior performance to cache-based methods, while delivering a 4$\times$ model compression and significant speedup improvement.
Furthermore, CacheQuant demonstrates robustness to cache frequency, as evidenced by its consistent outperformance in Table~\ref{tab:cifar}. 
At smaller cache frequency, our method even achieves lower FID score than the full-precision models. 
This is a common occurrence observed in prior works~\cite{liu2024enhanced,li2023q,liu2024dilatequant}, suggesting that the generated image quality is comparable to that produced by the full-precision models.
We demonstrate the superiority of CacheQuant over fast solvers by comparing it with the PLMS solver~\cite{liu2022pseudo}. As shown in Table~\ref{tab:coco}, using Stable Diffusion with 50 PLMS steps as a baseline, reducing the PLMS steps to 20 severely degrades performance. In contrast, our method maintains performance while achieving a 4$\times$ model compression and more than a 5$\times$ speedup.

\begin{table}[h]\scriptsize %\footnotesize %\scriptsize %\tiny %\small
    \centering
    \caption{Class-conditional generation quality on ImageNet using LDM-4 with UNet framework, employing 250 DDIM steps. }
    \label{tab:imagenet}
    \setlength{\tabcolsep}{1.4mm}
    \begin{tabular}{l | c c c c | c c }
  \toprule
  \multicolumn{7}{c}{\centering \bf ImageNet 256 $\times$ 256}  \\
  \bf Method & \bf Bops $\downarrow$ & \bf Speed $\uparrow$ & \bf Size $\downarrow$ & \bf Retrain & \bf FID $\downarrow$ & \bf IS $\uparrow$ \\
  \midrule
  LDM-4 & 102.22T & 1.00$\times$ & 1824.6 & \xmark & 3.37 & 204.56 \\
  \midrule
  EDA-DM (W8A8) & 6.39T & 1.91$\times$ & 457.1 & \cmark & 4.13 & 186.78 \\
  EDA-DM (W4A8) & 3.19T & 1.91$\times$ & 229.2 & \cmark & 4.79 & 176.43 \\
  EDA-DM (W4A4) & 1.61T & 3.35$\times$ & 229.2 & \cmark & 44.12 & 62.04 \\
  Diff-Pruning & 53.98T & 1.51$\times$ & 757.7 & \cmark & 9.27 & 214.42 \\  
  \midrule
  Deepcache-N=5 & 24.06T & 4.12$\times$ & 1824.6 & \xmark & 3.79 & 199.58 \\
  ours-N=5 (W8A8) & 1.50T & 7.87$\times$ & 457.1 & \xmark & 4.03 & 193.90 \\
  ours-N=5 (W4A8) & 0.75T & 7.87$\times$ & 229.2 & \cmark & 6.26 & 168.46 \\
  \midrule
  Deepcache-N=10 & 14.31T & 6.96$\times$ & 1824.6 & \xmark & 4.60 & 188.81 \\
  ours-N=10 (W8A8) & 0.89T & 12.20$\times$ & 457.1 & \xmark & 4.68 & 184.38 \\
  ours-N=10 (W4A8) & 0.45T & 12.20$\times$ & 229.2 & \cmark & 6.90 & 158.27 \\
  \midrule
  Deepcache-N=15 & 11.17T & 9.19$\times$ & 1824.6 & \xmark & 5.91 & 175.50 \\
  ours-N=15 (W8A8) & 0.70T & 16.55$\times$ & 457.1 & \xmark & 5.51 & 174.81 \\
  ours-N=15 (W4A8) & 0.35T & 16.55$\times$ & 229.2 & \cmark & 9.40 & 139.64 \\
  \midrule
  Deepcache-N=20 & 9.62T & 10.54$\times$ & 1824.6 & \xmark & 8.08 & 159.27 \\
  ours-N=20 (W8A8) & 0.60T & 18.06$\times$ & 457.1 & \xmark & 7.21 & 160.68 \\
  ours-N=20 (W4A8) & 0.30T & 18.06$\times$ & 229.2 & \cmark & 12.65 & 124.13 \\
  \bottomrule
\end{tabular}
    \vspace{-0.2cm}
\end{table}

\begin{table}[h]\scriptsize %\footnotesize %\scriptsize %\tiny %\small
    \centering
    \caption{Class-conditional generation quality on ImageNet using DiT-XL/2 with DiT framework, employing 50 DDIM steps.}
    \label{tab:imagenet1}
    \setlength{\tabcolsep}{1.4mm}
    \begin{tabular}{l | c c c c | c c }
  \toprule
  \multicolumn{7}{c}{\centering \bf ImageNet 256 $\times$ 256}  \\
  \bf Method & \bf Bops $\downarrow$ & \bf Speed $\uparrow$ & \bf Size $\downarrow$ & \bf Retrain & \bf FID $\downarrow$ & \bf IS $\uparrow$ \\
  \midrule
  DiT-XL/2 & 117.18T & 1.00$\times$ & 2575.42 & \xmark & 6.02 & 246.24 \\
  % \midrule
  % PTQ4DiT (W8A8) & 7.32T & 2.10$\times$ & 645.72 & \cmark & 5.45 & 250.68 \\
  % PTQ4DiT (W4A8) & 3.67T & 2.10$\times$ & 323.79 & \cmark & 9.17 & 179.95 \\
  \midrule
  $\Delta$-DiT-N=2 & 87.88T & 1.31$\times$ & 2575.42 & \xmark & 9.06 & 205.95 \\
  ours-N=2 (W8A8) & 5.49T & 2.72$\times$ & 645.72 & \xmark & 7.86 & 213.08 \\
  \midrule
  $\Delta$-DiT-N=3 & 75.53T & 1.51$\times$ & 2575.42 & \xmark & 13.75 & 171.68 \\
  ours-N=3 (W8A8) & 4.72T & 3.08$\times$ & 645.72 & \xmark & 12.42 & 173.17 \\
  \bottomrule
\end{tabular}
    \vspace{-0.3cm}
\end{table}

\begin{table*}[t]\footnotesize %\footnotesize %\scriptsize %\tiny %\small
    \centering
    \caption{Text-conditional generation quality on PartiPrompt and MS-COCO using Stable Diffusion with UNet framework.}
    \label{tab:coco}
    \setlength{\tabcolsep}{1.8mm}

\begin{tabular}{l | c c c c | c c c c | c c}
  \toprule
  & & & & & \multicolumn{4}{c|}{\bf MS-COCO} & \multicolumn{2}{c}{\bf PartiPrompts} \\
  \bf Method & \bf Prec.(W/A) & \bf Bops $\downarrow$ & \bf Size $\downarrow$ & \bf Retrain & \bf Speed $\uparrow$ & \bf FID $\downarrow$ & \bf IS $\uparrow$ & \bf CLIP Score $\uparrow$ & \bf Speed $\uparrow$ & \bf CLIP Score $\uparrow$ \\
  \midrule
  PLMS - 50 steps & 32/32 & 346.96T & 4112.5 & \xmark & 1.00$\times$ & 25.59 & 41.02 & 26.89 & 1.00$\times$ & 27.23 \\
  \midrule
  PLMS - 20 steps & 32/32 & 138.78T & 4112.5 & \xmark & 2.43$\times$ & 24.70 & 39.72 & 26.74 & 2.46$\times$ & 27.04 \\
  Deepcache - N=10 & 32/32 & 133.58T & 4112.5 & \xmark & 3.52$\times$ & 23.45 & 39.21 & 26.71 & 3.56$\times$ & 26.86 \\
  Small SD & 16/16 & 57.28T & 1158.6 & \cmark & 2.93$\times$ & 29.43 & 32.55 & 26.02 & 2.80$\times$ & 25.99 \\
  BK-SDM - Base & 16/16 & 57.30T & 1160.2 & \cmark & 2.79$\times$ & 28.47 & 36.79 & 26.21 & 2.66$\times$ & 26.53 \\
  BK-SDM - Small & 16/16 & 55.75T & 966.6 & \cmark & 2.88$\times$ & 30.10 & 35.57 & 25.22 & 2.80$\times$ & 25.76 \\
  BK-SDM - Tiny & 16/16 & 52.50T & 648.5 & \cmark & 2.92$\times$ & 31.82 & 32.82 & 25.10 & 2.87$\times$ & 25.51 \\
  \rowcolor[gray]{0.9} Ours - N=5 & 8/8 & 8.44T & 1029.7 & \xmark & \bf 5.18$\times$ & 23.74 & \bf 39.81 & \bf 26.87 & \bf 5.20$\times$ & 27.13 \\
  \rowcolor[gray]{0.9} Ours - N=5 & 4/8 & \bf 4.27T & \bf 515.9 & \cmark & \bf 5.18$\times$ & \bf 23.23 & 39.41 & 26.77 & \bf 5.20$\times$ & \bf 27.15 \\
  \bottomrule
\end{tabular}
    \vspace{-0.3cm}
\end{table*}

\subsection{Comparison with Structural-level Methods}
The structural-level acceleration methods primarily include model quantization, pruning, and distillation.
We compare CacheQuant with quantization-based methods (EDA-DM~\cite{liu2024enhanced}) in Table~\ref{tab:imagenet}. At 8-bit precision, CacheQuant with $N$=5 cache frequency outperforms EDA-DM (FID 4.03 vs 4.13). Importantly, CacheQuant avoids costly retraining and achieves significant acceleration improvements (Speed 7.87$\times$ vs 1.91$\times$).
As the bit width decreases, EDA-DM with the 4-bit precision achieves an 8$\times$ compression and a 3.35$\times$ speedup. However, the FID score significantly drops to 44.12.
In stark contrast, CacheQuant combined with reconstruction maintains an FID score of 12.65, achieving 8$\times$ compression and 18.06$\times$ speedup.
We conduct a comparison with pruning-based method in Table~\ref{tab:imagenet}. As can be seen, CacheQuant surpasses Diff-Pruning~\cite{fang2023structural} in terms of efficiency, performance, acceleration, and compression.
We also compare CacheQuant with distillation-based methods, including Small SD~\cite{Lu2022KnowledgeDO} and BK-SDM~\cite{kim2023bk}, which are developed by retraining on LAION~\cite{schuhmann2021laion} dataset, using Stable Diffusion as the baseline.
As reported in Table~\ref{tab:coco}, our method achieves superior performance and faster acceleration compared to these approaches. 

\subsection{Analysis}
\paragraph{Ablation Study.}
To assess the efficacy of each proposed component, we conduct a comprehensive ablation study on ImageNet, employing the LDM-4 model with 250 steps, as presented in Table~\ref{tab:ablation}.
We add 8-bit quantization to DeepCache with $N$=20 cache frequency as a baseline, resulting in an increase of the FID score to 15.36.
With DPS introduced to select the optimal cache schedule, the FID score significantly improves to 8.47. This demonstrates that DPS effectively minimizes errors caused by caching and quantization.
By further incorporating EDC that corrects decoupled errors in a training-free manner, the FID score is improved to 7.21.
Moreover, our method, combined with reconstruction approach, further enhances performance, notably increasing the IS score to 180.42.

\begin{table}[h]\footnotesize %\footnotesize %\scriptsize %\tiny %\small
    \centering
    \caption{The effect of different components proposed in the paper. }
    \label{tab:ablation}
    \setlength{\tabcolsep}{1.5mm}
    \begin{tabular}{ccccc}
        \toprule 
        \bf Method & \bf Prec.(W/A) & \bf Retrain & \bf FID $\downarrow$ & \bf IS $\uparrow$ \\ 
        \cmidrule(r){1-5}
        Deepcache-N=20 & 32/32 & \xmark & 8.08 & 159.27 \\ 
        baseline & 8/8 & \xmark & 15.36 & 121.78 \\ 
        +DPS & 8/8 & \xmark & 8.47 & 154.07 \\ 
        +DPS+EDC & 8/8 & \xmark & 7.21 & 160.68 \\
        +DPS+EDC+Recon & 8/8 & \cmark & 6.34 & 180.42 \\
        \bottomrule
    \end{tabular}
    \vspace{-0.3cm}
\end{table}

\begin{figure}[h]
    \centering
    \includegraphics[width=1.0\linewidth]{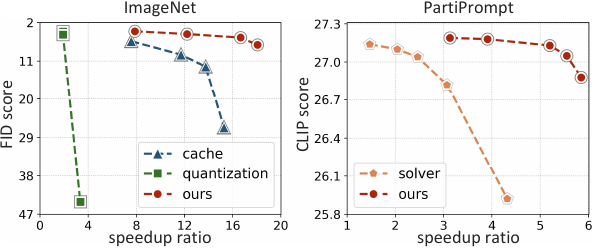}
    \caption{An overview of the acceleration-vs-performance tradeoff across various approaches. Data from LDM-4 on ImageNet and Stable Diffusion on PartiPrompt.}
    \label{fig:tradeoff}
    \vspace{-0.3cm}
\end{figure}

\vspace{-0.3cm}
\paragraph{Acceleration vs. Performance Tradeoff.}\label{sec:tardeoff}
We investigate the tradeoff between acceleration and performance for various approaches, as presented in Figure~\ref{fig:tradeoff}.
As speedup ratio increases, traditional acceleration methods, such as cache (Deepcache), quantization (EDA-DM), and solvers (PLMS), suffer from significant performance degradation.
In sharp contrast, our method comprehensively accelerates diffusion models at two levels, further pushing acceleration limits while maintaining performance.
The detail experimental settings are reported in Appendix~\ref{app:tradeoff}.

\begin{figure}[h]
    \centering
    \includegraphics[width=1.0\linewidth]{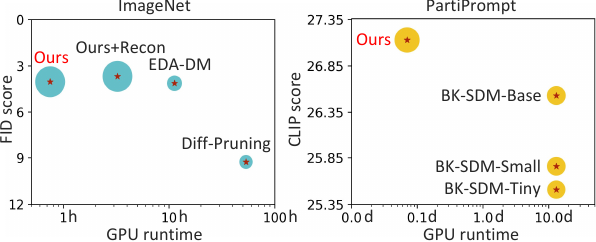}
    \caption{Comparison of the efficiency across various approaches. Data from LDM-4 on ImageNet and Stable Diffusion on PartiPrompt. The circle size denotes speedup ratio.}
    \label{fig:efficiency}
    \vspace{-0.3cm}
\end{figure}

\vspace{-0.3cm}
\paragraph{Study on Efficiency.}
As shown in Figure~\ref{fig:efficiency}, our method significantly outperforms traditional approaches in efficiency. For instance, compression-based methods require over 10 hours of GPU runtime, while distillation-based methods demand more than 10 days to complete. 

\begin{figure}[h]
    \centering
    \includegraphics[width=1.0\linewidth]{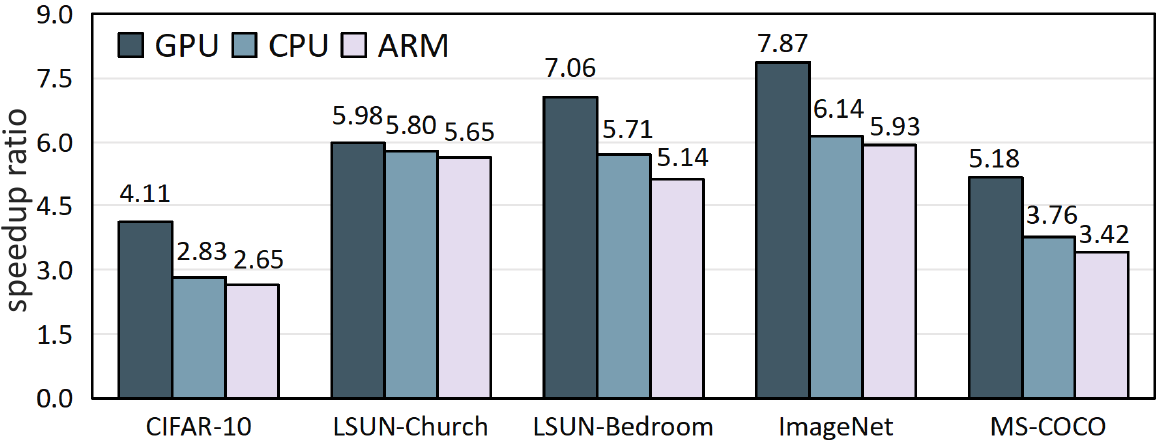}
    \caption{Speedup ratio of diffusion models with 8-bit precision and $N$=5 cache frequency.}
    \label{fig:acc}
    \vspace{-0.3cm}
\end{figure}

\vspace{-0.3cm}
\paragraph{Deployment of accelerated models.}
To evaluate the real-world speedup, we deploy our accelerated diffusion models on various hardware platforms.
As shown in Figure~\ref{fig:acc}, the acceleration on GPU is significantly more pronounced compared to CPU and ARM. Our method achieves a 5$\times$ GPU speedup of Stable Diffusion on MS-COCO, significantly facilitating its applications in real-world scenarios.

\section{Conclusion}
In this paper, we introduce CacheQuant, a novel training-free paradigm that comprehensively accelerates diffusion models at both temporal and structural levels.
To address the non-orthogonality of optimization, we propose DPS that selects the optimal cache schedule to minimize errors caused by caching and quantization.
Additionally, we employ DEC to further mitigate the coupled and accumulated errors without any retraining.
Empirical evaluations on several datasets and different model frameworks demonstrate that CacheQuant outperforms traditional acceleration methods. 
Importantly, the proposed paradigm pushes the boundaries of diffusion model acceleration while maintaining performance, thereby offering a new perspective in the field. 

\section{Acknowledge}
This work is supported in part by the National Science and Technology Major Project of China under Grant 2022ZD0119402; in part by the National Natural Science Foundation of China under Grant 62276255.

{
    \small
    \bibliographystyle{ieeenat_fullname}
    \bibliography{main}
}

\clearpage
\setcounter{page}{1}
\maketitlesupplementary

\section{Detailed experimental implementations}\label{app:detail}
We use the pre-training models of DiT-XL/2\footnote{\url{https://github.com/facebookresearch/DiT}}, DDIMs\footnote{\url{https://github.com/ermongroup/ddim}}, and LDMs\footnote{\url{https://github.com/CompVis/latent-diffusion}} from the official website.
For Stable Diffusion, we use the CompVis codebase\footnote{\url{https://github.com/CompVis/stable-diffusion}} and its v1.4 checkpoint.
The conditional generation models consist of a diffusion model and a decoder model.
Like the previous works~\cite{li2023q,wu2024ptq4dit,liu2024enhanced}, we focus only on the diffusion model and does not quantize the decoder model.
In the reconstruction training, we set the calibration samples to 1024 and the training batch to 32 for all experiments. However, for the Stable Diffusion, we adjust the reconstruction calibration samples to 512 and the training batch to 4 due to time and memory source constraints.
We use open-source tool \emph{pytorch-OpCounter}\footnote{\url{https://github.com/Lyken17/pytorch-OpCounter}} to calculate the Size and Bops of models before and after quantization.
And following the quantization setting, we only calculate the diffusion model part, not the decoder and encoder parts.
We use the ADM’s TensorFlow evaluation suite \emph{guided-diffusion}\footnote{\url{https://github.com/openai/guided-diffusion}} to evaluate FID and IS, and use the open-source code \emph{clip-score}\footnote{\url{https://github.com/Taited/clip-score}} to evaluate CLIP scores. 
The accelerated diffusion models are deployed by utilizing CUTLASS\footnote{\url{https://github.com/NVIDIA/cutlass}} and PyTorch\footnote{\url{https://pytorch.org/blog/quantization-in-practice/}}.
The speed up ratio is calculated by measuring the time taken to generate a single image on the RTX 3090.
As per the standard practice~\cite{nichol2021glide,liu2024enhanced,liu2024dilatequant}, we employ the zero-shot approach to evaluate Stable Diffusion on COCO-val, resizing the generated 512 $\times $ 512 images and validation images in 300 $\times $ 300 with the center cropping to evaluate FID and IS score.

\section{Express $X_{cq} W_q$ with two correction methods}\label{app:Xcq}
Based on Eq.~\ref{eq:O} and Eq.~\ref{eq:cor}, the direct correction simply expresses $X_{cq} W_q$ as:
\begin{align}
    X_{cq} W_q &= \frac{X_g W_g}{a} - \frac{b}{a} 
\end{align}

Our method corrects for $X_c$ and correct $O_{cq}$, respectively. 
Based on Eq.~\ref{eq:O} and Eq.~\ref{eq:cor}, derive the equation: 
\begin{align}
    X_c &= \frac{X_g}{a_1} - \frac{b_1}{a_1} \\
    X_{cq} W_q &= \frac{X_c W_g}{a_2} - \frac{b_2}{a_2}
\end{align}
Furthermore, the expression for $X_{cq} W_q$ is as:
\begin{align}
    X_{cq} W_q &= \frac{X_g}{a_1} \cdot \frac{W_g}{a_2} - \frac{b_1}{a_1} \cdot \frac{W_g}{a_2} - \frac{b_2}{a_2}
\end{align}
Since the correction parameters $(a, b) \in \mathbb{R}^{C^o}$ and $(a_1, b_1) \in \mathbb{R}^{C^i}$, $(a_2, b_2) \in \mathbb{R}^{C^o}$, the two representations of $X_{cq} W_q$ are equivalent if and only if $a_1 = \mathbf{1}$ and $b_1 = \mathbf{0}$.
\begin{table}[t]\footnotesize %\scriptsize %\footnotesize %\tiny %\small
    \centering
    \caption{Results of LDM-4 on ImageNet with 250 DDIM steps.}
    \label{tab:tradeoff1}
    \setlength{\tabcolsep}{1.10mm}
    \begin{tabular}{c l | c c c }
      \toprule
      \multicolumn{2}{c}{\centering \bf Method} & \bf Retrain & \bf Speed $\uparrow$ & \bf FID $\downarrow$ \\
      \midrule
      \multirow{4}{1.5cm}{\centering Cache} & Deepcache-$N$=12 & \xmark & 7.58$\times$ & 6.35 \\
      \multirow{4}{*}{} & Deepcache-$N$=25 & \xmark & 11.71$\times$ & 9.51 \\
      \multirow{4}{*}{} & Deepcache-$N$=35 & \xmark & 13.75$\times$ & 12.32 \\
      \multirow{4}{*}{} & Deepcache-$N$=50 & \xmark & 15.28$\times$ & 26.63 \\
      \midrule
      \multirow{3}{1.5cm}{\centering Quantization} & EDA-DM (W8A8) & \cmark & 1.91$\times$ & 4.13 \\
      \multirow{3}{*}{} & EDA-DM (W4A8) & \cmark & 1.91$\times$ & 4.13 \\
      \multirow{3}{*}{} & EDA-DM (W4A4) & \cmark & 3.35$\times$ & 44.12 \\
      \midrule
      \multirow{4}{1.5cm}{\centering Ours} & CacheQuant-$N$=5 (W8A8) & \xmark & 7.87$\times$ & 4.03 \\
      \multirow{4}{*}{} & CacheQuant-$N$=10 (W8A8) & \xmark & 12.20$\times$ & 4.68 \\
      \multirow{4}{*}{} & CacheQuant-$N$=15 (W8A8) & \xmark & 16.65$\times$ & 5.51 \\
      \multirow{4}{*}{} & CacheQuant-$N$=20 (W8A8) & \xmark & 18.06$\times$ & 7.21 \\
      \bottomrule
    \end{tabular}
\end{table}

\begin{table}[t]\footnotesize %\scriptsize %\footnotesize %\tiny %\small
    \centering
    \caption{Results of Stable Diffusion on PartiPrompt with 50 PLMS steps.}
    \label{tab:tradeoff2}
    \setlength{\tabcolsep}{1.10mm}
    \begin{tabular}{c l | c c c }
      \toprule
      \multicolumn{2}{c}{\centering \bf Method} & \bf Retrain & \bf Speed $\uparrow$ & \bf CLIP score $\uparrow$ \\
      \midrule
      \multirow{5}{0.8cm}{\centering PLMS} & PLMS-35 steps & \xmark & 1.47$\times$ & 27.14 \\
      \multirow{5}{*}{} & PLMS-25 steps & \xmark & 2.04$\times$ & 27.10 \\
      \multirow{5}{*}{} & PLMS-20 steps & \xmark & 2.46$\times$ & 27.04 \\
      \multirow{5}{*}{} & PLMS-15 steps & \xmark & 3.07$\times$ & 26.82 \\
      \multirow{5}{*}{} & PLMS-10 steps & \xmark & 4.32$\times$ & 25.92 \\
      \midrule
      \multirow{5}{0.8cm}{\centering Ours} & CacheQuant-$N$=2 (W8A8) & \xmark & 3.13$\times$ & 27.19 \\
      \multirow{5}{*}{} & CacheQuant-$N$=3 (W8A8) & \xmark & 3.91$\times$ & 27.18 \\
      \multirow{5}{*}{} & CacheQuant-$N$=5 (W8A8) & \xmark & 5.20$\times$ & 27.05 \\
      \multirow{5}{*}{} & CacheQuant-$N$=6 (W8A8) & \xmark & 5.55$\times$ & 27.05 \\
      \multirow{5}{*}{} & CacheQuant-$N$=8 (W8A8) & \xmark & 5.85$\times$ & 26.88 \\
      \bottomrule
    \end{tabular}
\end{table}
\section{Experimental settings for evaluation of acceleration-vs-performance tradeoff}\label{app:tradeoff}
We evaluate the tradeoff between acceleration and performance for various approaches in Sec~\ref{sec:tardeoff}.
The detail experimental settings and results in Figure~\ref{fig:tradeoff} are shown in Table~\ref{tab:tradeoff1} and~\ref{tab:tradeoff2}.

\begin{figure*}[t]
    \centering
    \includegraphics[width=1.0\linewidth]{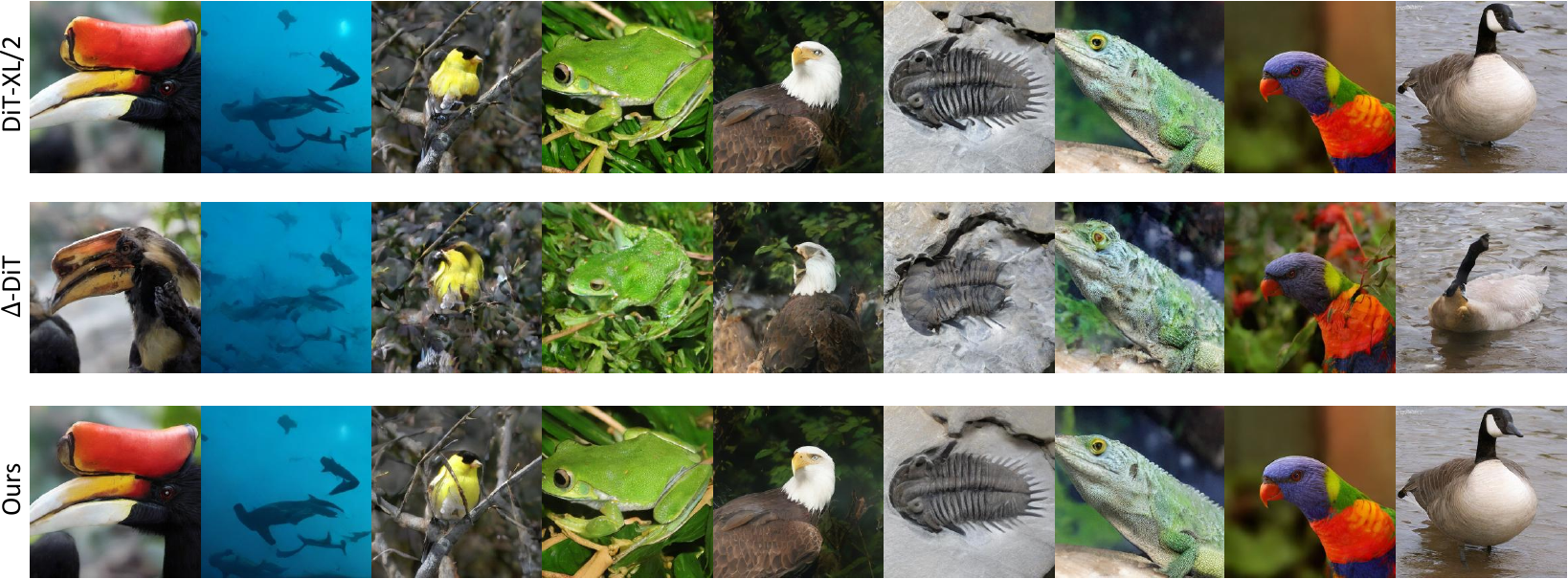}
    \caption{Visualization of the generated images by $\Delta$-DiT and CacheQuant, with $N$=2 cache frequency.}
    \label{fig:dit}
    \vspace{-0.3cm}
\end{figure*}

\begin{figure*}[t]
    \centering
    \includegraphics[width=1.0\linewidth]{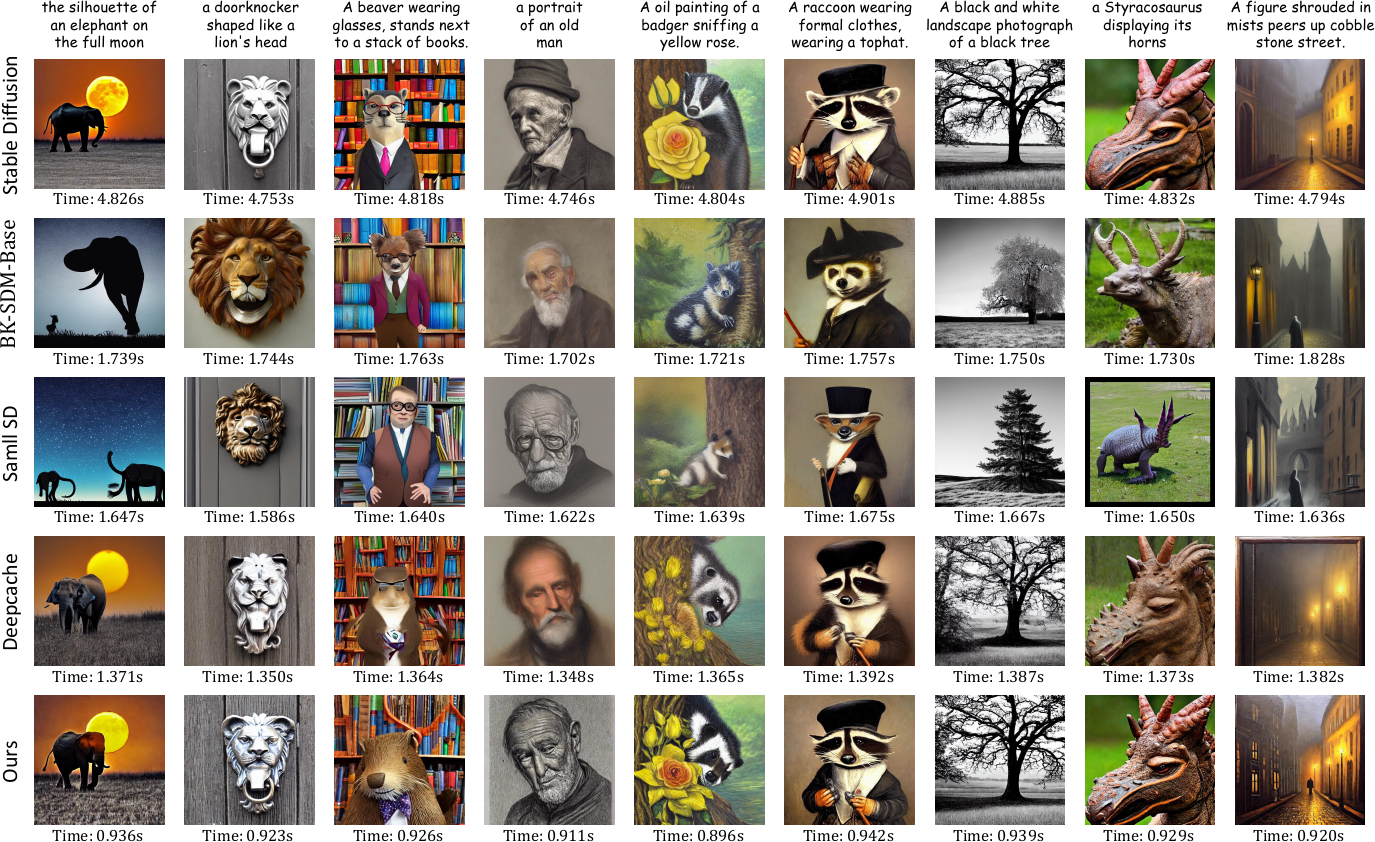}
    \caption{Visualization of the generated images by BK-SDM-Base, Small SD, Deepcache with $N$=10 cache frequency, and CacheQuant. All the methods adopt the 50-step PLMS. The time here is the duration to generate a single image.}
    \label{fig:sd}
    \vspace{-0.3cm}
\end{figure*}

\section{Comparison of generated results}\label{app:sample}
Within this section, we present random samples derived from original models and other accelerated methods with a fixed random seed. 
Our method maintains 8-bit precision.
We visualize the generated image quality and latency of different methods in Figures~\ref{fig:dit} and~\ref{fig:sd}.

\section{Limitations and future work}\label{app:limitation}
While CacheQuant achieves remarkable results in a training-free manner at 8-bit precision, it relies on reconstruction to recovery performance at W4A8 precision.
In the future, we will further refine CacheQuant to improve its compatibility with W4A8 precision.

\end{document}